\documentclass{article}
\usepackage{spconf,amsmath,graphicx}
\usepackage{xcolor}
\usepackage{float}
\usepackage{subfig}
\usepackage{multirow}
\usepackage{devanagari}
\usepackage{booktabs}
\usepackage{colortbl}
\usepackage{amsmath}
\usepackage{amssymb} 

\title{Balanced End-to-End Monolingual pre-training for Low-Resourced Indic Languages Code-Switching Speech Recognition}

\name{Amir Hussein$^{1,2}$, Shammur Chowdhury$^3$, Najim Dehak$^{2}$, Ahmed Ali$^3$}
\address{
  $^1$Kanari AI, California, USA \\
  $^2$Center for Language and Speech Processing,
  Johns Hopkins University, Baltimore, USA\\
  $^3$Qatar Computing Research Institute, HBKU, Doha, Qatar}

\begin{document}
\ninept
\maketitle

\begin{abstract}

The success in designing Code-Switching (CS) ASR often depends on the availability of the transcribed CS resources. Such dependency harms the development of ASR in low-resourced languages such as Bengali and Hindi. In this paper, we exploit the transfer learning approach to design End-to-End (E2E) CS ASR systems for the two low-resourced language pairs using different monolingual speech data and a small set of noisy CS data. We trained the CS-ASR, following two steps: (\textit{i}) building a robust bilingual ASR system using a convolution-augmented transformer (Conformer) based acoustic model and n-gram language model, and (\textit{ii}) fine-tuned the entire E2E ASR with limited noisy CS data. We tested our method on MUCS 2021 challenge  and achieved 3rd place in the CS track. 
Unlike, the leading two systems that benefited from crawling YouTube and learning transliteration pairs, our proposed transfer learning approach focused on using only the limited CS data with no data-cleaning or data re-segmentation. Our approach achieved 14.1\% relative gain in word error rate (WER) in Hindi-English and 27.1\% in Bengali-English. We provide detailed guidelines on the steps to finetune the self-attention based model for limited data for ASR. Moreover, we release the code and recipe used in this paper.
\end{abstract}


\section{Introduction}

The rise of globalization impacted our life in many ways, leading general acquisition of cross-lingual communication needs. With prevalent multilingualism, the voice technologies, such as automatic speech recognition (ASR), are expected to understand mixed language input and deal with code-switching (CS). 
CS is a common phenomena in a multicultural society, where speakers often alter between two or more languages. CS occurs in spontaneous speech -- formal and semi-formal settings like educational lectures and news \cite{sitaram2019survey}, are highly unpredictable and difficult to model. Building such CS-ASR is very challenging, mainly due to the scarcity of transcribed data and highly unbalanced language distribution.

There has been increasing interest in building such CS-ASR for handful of language pairs, such as -- Mandarin-English \cite{li2013improved}, Hindi-English \cite{sreeram2020exploration}, French-Arabic \cite{amazouz2017addressing}, Arabic-English \cite{ali2021arabic, hamed2021investigations} and English-Arabic-French \cite{chowdhury2021towards}.  
The end-to-end (E2E) systems have gained more popularity recently over conventional hybrid systems. E2E outperformed modular systems in modeling monolingual and multilingual systems \cite{toshniwal2018multilingual, datta2020language,chowdhury2021towards}. This can be owed to the fact that E2E optimizes all parts of the network for the overall word error rate (WER). Researchers in \cite{luo2018towards,shan2019investigating} proposed using additional language identification task on top of connectionist temporal classification (CTC) Attention (CTC-Attention) \cite{kim2017joint} architecture to detect the CS point in English-Mandarin speech. In \cite{sreeram2020exploration}, authors modeled limited Hindi-English CS using E2E attention with context-dependent target to word transduction, factorized language model with part-of-speech (POS) tagging and CS identification, and proposed textual features to enhance the context modeling in CS. In \cite{zhou2020multi}, authors proposed transformer-based architecture with two symmetric language-specific encoders to capture the individual language attributes for Mandarin-English CS, whereas in \cite{chowdhury2021towards}, the authors utilized multilingual strategy to model CS along with dialectal language varities. 


In this paper, we build on the aforementioned contributions to develop E2E speech recognition systems for low-resourced languages. To lessen the dependency on large data availability, we proposed a novel strategy exploiting transfer learning using monolingual datasets and a small amount of CS data. We utilize the monolingual datasets to build a robust bilingual ASR, and then we finetuned the model for CS phenomena with handful of error-prone noisy data. 
We evaluated our strategy using the small
Hindi-English and Bengali-English CS data, collected from technical tutorials,  released with the MUCS 2021 CS task \cite{diwan2021multilingual}.\footnote{https://navana-tech.github.io/MUCS2021/} 
In the spoken tutorials, the speakers use Hindi or Bengali as native languages and the English as the non-native/second language, thus creating frequent CS scenarios. Moreover, the dataset pairs are also very challenging due to the quality of the CS data along with the scarcity of publicly available resources. Hence, it is a perfect candidate to test our approach. 


The first places in the competition was achieved by \cite{diwan2021multilingual, wiesner2021training} which was based on the hybrid HMM-DNN system. The authors in \cite{wiesner2021training} derived non-standard pronunciations
by leveraging transliteration pairs, and acoustically driven pronunciation
modeling. The main limitation of \cite{wiesner2021training} is that it highly depends on the in-domain knowledge and hard to generalize on out-of domain or different language-pairs. 
On the other hand, authors in \cite{diwan2021multilingual} crawled around 1,000 hours of YouTube videos from similar domain using semi-supervised approach, which is great in cases where data is publicly available. 

We propose a transfer learning approach with E2E conformer model that utilizes only publicly available limited monolingual data and does not require any domain knowledge. The proposed approach achieves significant improvements in CS task with two steps: (\textit{i}) Balanced pre-training on monolingual languages, and (\textit{ii}) Careful fine-tuning on the CS data. In our approach, the acoustic model was built based on the recently introduced end-to-end (E2E) convolution-augmented transformer (conformer) for speech recognition \cite{conformer}. 
In \cite{guo2020recent}, authors showed that the conformer significantly outperforms the traditional transformer in most of the ASR tasks. For language modeling (LM), we used word level bi-gram  model as it provided best results on limited CS data compared to other deep learning methods. To train the LM models, we used publicly available monolingual data in addition to the CS data. In summary, the key contributions of our work include:
\begin{itemize}
    \item A balanced monolingual transfer learning approach for low resourced Indic CS data.
    \item A detailed practical guidelines for the E2E conformer pre-training and fine-tuning strategy with limited CS data.
\end{itemize}

\noindent We release the code and recipe  used for this paper for further research.\footnote{https://github.com/AmirHussein96/IS21-CS-E2E}

The rest of this paper is organized as follows:
Section \ref{arch} presents the architecture used to development of the acoustic model and the language modeling. Section \ref{dataset} presents datasets description.  The details of the experimental setup, results and their discussion are given in Section \ref{experiments}. Section \ref{conclusion} concludes the findings of our study.
\section{Architecture}\label{arch}
\subsection{Acoustic Modeling}\label{AM}
We develop ASR conformer architecture using ESPNET toolkit \cite{guo2020recent}. The implementation consists of a conformer encoder \cite{conformer} which is a multi-blocked architecture and a transformer decoder. The encoder consists of several blocks, each is a stack of a position-wise feed-forward (FF) module, multi-head self-attention (MHSA), a convolution operation (CONV) module, and another FF module in the end. The self-attention computation of every single head in MHSA can be formulated as:
\begin{equation}\label{eq1}
\begin{small}
Att_h(\mathbf{Q_h,K_h,V_h}) = S(\dfrac{\mathbf{Q_h} * \mathbf{K_h}^T}{\sqrt{d^k}})*\mathbf{V_h} 
\end{small}
\end{equation}
where $S$ is a softmax operation, $\mathbf{Q} = \mathbf{X} * \mathbf{W^q}$, $\mathbf{K} = \mathbf{X} * \mathbf{W^{k}}$, and $\mathbf{V} = \mathbf{X} * \mathbf{W^{v}}$ are the queries, keys and values respectively. The $\mathbf{W^{q}}$ and $\mathbf{W^{k}}$ $\in$  $\mathbb{R}^{d^{att}\times d^{\mathrm{k}}}$ and $\mathbf{W^{v}}$ $\in$ $\mathbb{R}^{d^{att}\times  
d^{\mathrm{v}}}$ are learnable weights. The $d^{att}$ is the dimension of the attention, and $d^{\mathrm{v}}$, $d^{\mathrm{k}} = d^{\mathrm{q}}$ are the dimensions of values, keys and queries. To simultaneously attend to information from different representations, outputs of each head are concatenated in MHSA.
The MHSA is followed by a convolution module (CONV) which consists of a 1-D convolution layer, gated linear
units (GLU) activation batch normalization (BN) layer, and a Swish activation. 
Each module includes layer normalization (LN) and is followed by a layer dropout (D), and a residual connection from the  module input.

\subsubsection{Conformer ASR training}

During the training, the conformer ASR predicts the target sequence $\mathbf{Y}$ of tokens from acoustic features $\mathbf{X}$. For text tokenization, we used word-piece byte-pair-encoding (BPE)~\cite{kudo2018sentencepiece}. The total loss function $\mathcal{L}_{asr}$ is a multi-task learning objective that combines the decoder cross-entropy (CE) loss $\mathcal{L}_{\mathrm{ce}}$ and the CTC loss \cite{graves2006connectionist} $\mathcal{L}_{\mathrm{ctc}}$.
\begin{equation}\label{eq6}
\mathcal{L}_{asr}=\alpha \mathcal{L}_{\mathrm{ctc}}+(1-\alpha) \mathcal{L}_{\mathrm{ce}}
\end{equation}
where $\alpha$ is a weighting factor with the selected best value of 0.3. In our approach, the conformer is first pre-trained with monolingual speech data from both Hindi/Bangali and English with shared vocabulary for both languages. We add around half of the available CS data to make the model familiar with CS examples that mix the two languages. Then, we fine-tune all the model parameters on all the available CS speech with a very small learning rate ($\frac{1}{50}$ of lr used during the pre-training).  

\subsection{Language Modeling}\label{LM}
In practical scenarios for low-resourced languages, the availability of 
CS text data is very limited. Hence, we decided to train word-level n-gram language models (LMs): a 2-gram and a 3-gram LMs. Both n-gram models were trained with the KenLM toolkit \cite{heafield2011kenlm} on the entire text data described in Section \ref{dataset}.
During the decoding, the best transcription is selected by leveraging both the posteriors of an acoustic model (AM) and the perplexity of a language model (LM).

\section{Datasets Description}\label{dataset}

In this section, we describe the details of the provided MUCS 2021 data for Code-switching subtask-2 \cite{diwan2021multilingual}. In addition, we also present the publicly available acoustic and text resources that are used in developing our approach.

\subsection{MUCS21 Speech data}

The code-switching challenge used Hindi-English and Bengali-English datasets recorded from spoken tutorials covering various technical topics with the following challenges:

\begin{enumerate}
\item Misalignment between the transcription and segment start and end times.
\item Inconsistency in the script used to write the same word (some English words were written in the Latin script and some in the native scripts of Hindi and Bengali).
\item Some English words are merged with the native Hindi/Bengali words as one word.
\item In some cases, the transcription of the spoken utterance was found inaccurate or completely wrong.
\item Incomplete audio due to segmentation issue.
\end{enumerate}

The datasets are sampled at 16 kHz with 16 bits encoding. Basic analysis showed that each dataset contains around 45\% of non-native words and 55\% of Hindi/Bengali native words. 

\subsection{Publicly available Speech data}
In addition to the provided CS speech datasets, we used publicly available monolingual -- Bengali (Bn) \cite{kjartansson-etal-sltu2018} dataset, Hindi (Hi)\footnote{https://navana-tech.github.io/IS21SS-indicASRchallenge/data.html}  speech from the MUCS 2021 multilingual challenge, and Tedlium3 \cite{hernandez2018ted}. All the speech data are sampled at 16kHz except Hi which was sampled at 8kHz. As a result, we upsampled the Hi audio to 16kHz. Since each Hindi and Bengali datasets are limited, we use different subsets of Tedlium3\footnote{https://openslr.magicdatatech.com/51/} ranging from 22.7 hours to 203 hours to show the effect of selecting balanced data. More details about the datasets are shown in Table \ref{table1:hi_bn_cs}.

\begin{table}[!ht]
    \caption{MUCS 2021 challenge code-switching  and monolingual speech datasets.}
    \label{table1:hi_bn_cs}
    \scalebox{0.8}{
    \begin{tabular}{lrrrr}
      \toprule %
      \textbf{Dataset} & \textbf{Type} & \textbf{Hours} & \textbf{\#Segments} & \textbf{Vocab size}     \\
      \midrule
      \textbf{Hi-En}  & Train & 89.86  & 52,823 & 17,877   \\ 
       
        & Dev & 5.18  & 3136 & - \\
        & Hidden & 6.24 & 4,034 & - \\
        \midrule
        \textbf{Bn-En} & Train & 46.11  & 26,606 & 13,656 \\
         & Dev & 7.02  & 4,275 & - \\
         & Hidden & 5.53 & 3,130 & - \\
         \midrule
         \textbf{Hi}  & Train & 95.05  & 99925 &  6,542 \\ 
       
        & Dev & 2.6  & 1,950  & - \\
        \midrule
        \textbf{Bn} & Train & 211.6  & 214,703 & 27,607 \\
         & Dev & 2.6 & 2,700 & - \\
         \midrule
         \textbf{Tedlium$3$} & Train & 22.7 - 203.5  & 34,311 - 120,963  & 21,909 - 39,703 \\
         & Dev & 2.6  & 1,155 & -\\
      \bottomrule
      
    \end{tabular}
    }
    \vspace{-0.3cm}
\end{table}

\subsection{Text data}
For CS language modeling, we used Hindi-English news paper dataset,\footnote{https://www.kaggle.com/pk13055/code-mixed-hindienglish-dataset} Hindi Wikipedia articles\footnote{https://www.kaggle.com/disisbig/hindi-wikipedia-articles-172k}
and Bengali-English wiki dataset.\footnote{https://www.kaggle.com/abyaadrafid/bnwiki} 
In addition, to add conversational text, we utilize Tedlium3 transcription text along with the challenge CS transcription data.

\section{Experiments and Results}\label{experiments}

\subsection{Experimental Environment}
We ran our experiments on an HPC node equipped with 4 NVIDIA Tesla V100 GPUs with 16 GB memory, and 20 cores of Xeon(R) E5-2690 CPU.
\subsection{Data Processing}
We first augmented the raw speech data with the speed perturbation with speed factors of $0.9$, $1.0$, and $1.1$ \cite{ko2015audio}. Then, we extracted $83$-dimensional feature frames consisting of $80$-dimensional log Mel-spectrogram and pitch features \cite{ghahremani2014pitch} and applied cepstral mean and variance normalization (CMVN). Furthermore, we augmented these features using the specaugment approach \cite{park2019specaugment}. 
To reduce the noise in the provided CS transcription, we performed basic cleaning by removing all punctuation except the symbols that were spoken in the audio \{\_, /, =, +, \%, @\}. In addition, we converted English words to lowercase and separated the numbers and different words that were glued in one word: (attributes{\dn aTribyuTa} ==$>$ attributes {\dn aTribyuTa}).

\subsection{Default model hyperparameters}\label{hyperparameter_tuning}
All hyperparameters were obtained using a grid search during the pre-training phase. The E2E conformer-based ASR model was trained using Noam optimizer \cite{vaswani2017attention}.
Table (\ref{hyp_ASR_trans}) summarizes the best set of parameters that were found for the 
conformer architecture. Both models were pre-trained for 60 epochs with dropout-rate $0.1$, warmup-steps of $20000$, batch size of $64$ and learning rate of $5$. 

\begin{table}[!ht]
\centering
\caption{Values of E2E conformer hyperparameters obtained from the grid search. CNN: refers to CNN module kernel, Att: attention, Enc: encoder, Dec: decoder, and FF: fully connected layer }
\label{hyp_ASR_trans}
\resizebox{0.48\textwidth}{!}{
\begin{tabular}{|c|c|c|c|c|c|c|c|}
\hline
 {\textbf{Parameters}}& BPE & Att heads & CNN & Enc layers& Dec layers & $d^{k}$ & FF units  \\ \hline
 \textbf{Values} & $1000$ & $4$ & $15$ & $8$ & $4$ & $512$ & $2048$\\ 
\hline
\end{tabular}}
\vspace{-0.3cm}
\end{table}

\subsection{Pre-trained Models}
During the pre-training, the number of selected hours of non-native Tedlium3 was limited by the number of available hours of each native Hindi and Bengali data to avoid data biases. We used two configurations: (\textit{i}) Equal non-native (Ev), where the percentage of the non-native data is $45\%$ and the native data is $55\%$ (similar to the percentage of each language in the CS data); (\textit{ii}) Small non-native (Sv): the ratio of non-native to the native data is $1:4$. In addition, we added around half of the provided CS data during the pre-training. The number of selected hours for each configuration is summarized in Table \ref{table4:pre-training}.
\begin{table}[!ht]
 
  \begin{center}
    \caption{Number of hours of the monolingual (Hindi/Bengali), Tedlium3, and the CS used in pre-training phase for Ev and Sv configurations.}
    \label{table4:pre-training}
    \resizebox{0.3\textwidth}{!}{\begin{tabular}{llll}
      \toprule %
      \textbf{Configuration} & \textbf{Native} & \textbf{Tedlium3}  & \textbf{CS}    \\
      \midrule
      \textbf{Sv (Hindi)}    & 95 & 22.7 & 50 \\ 
      \midrule
       \textbf{Ev (Hindi)}    & 95 & 86 &  50 \\ 
        \midrule
        \textbf{Sv (Bengali)}    & 211.6 & 57.6 & 20.5\\
         \midrule
         \textbf{Ev (Bengali)}   & 211.6  & 200.5 & 20.5  \\
      \bottomrule
    \end{tabular}}
  \end{center}
  \vspace{-0.3cm}
\end{table}
\subsection{Fine-tuned Model}
 We adapted the pre-trained model for the CS task. For this we utilized 5.18 hours and 7.02 hours of Hi-En and Bn-En CS data provided in the competition for fine-tuning. Unlike previous studies that indicates advantages of freezing the part of the network, our empirical experiments shows that fine-tuning the entire network results more robust CS-ASR model. Using the same hyperparameters search space, the best learning rate for adaptation was found to be comparatively low (0.1), with respect to pre-training step. This is in aligned with many previous literature on transfer learning.

 


\subsection{Results}
We first present the results obtained from bilingual (pre-trained model). 
From manual enquiry, we observed that due to the noise levels in the evaluation set, reported WER is high even when the model produced better quality transcription. To select the best pre-trained model, we created our own development set consisting of 2.6 hours of each Hindi and Bengali monolingual sets and 2.6 hours of CS evaluation set. 
WERs in Table \ref{table5:pre-training}, indicates that the pre-trained models with only half of the provided CS data significantly outperformed the challenge baselines on average by 15\% in relative WER on CS development set. It is worth noting that the Hi-En baseline on the development set is not well representing the transcription quality. In fact, the automated transcription are of higher quality than the references.
\begin{table}[!htb]
  \begin{center}
    \caption{WER results on the locally created development set (l-Dev) from 2.6 hours of native Hindi (Hi) and Bengali (Bn), non-native Tedlium3 (Ted3), and the code-switching development set (CS-Dev).}
    \label{table5:pre-training}
    \resizebox{0.4\textwidth}{!}{\begin{tabular}{lllll}
      \toprule %
      \textbf{Configuration} & \textbf{Native} & \textbf{Ted3}  & \textbf{CS-Dev} & \textbf{l-Dev}   \\
      \midrule
       \textbf{Baseline (Hi-En) \cite{diwan2021multilingual}}  &  -  & - & 27.7 &  -
       \\
      \textbf{Baseline (Bn-En) \cite{diwan2021multilingual}} &  -  & - & 37.2 &  -
       \\
      \midrule
      
      \textbf{Sv (Hi-En)}  & \textbf{33.1}  & 19.6 & 28.3 &  26.9\\ 
 
       \textbf{Ev (Hi-En)}  & 35.1 & 12.7 & \textbf{27.5} & 25.4 \\ 
       \textbf{Ev (Hi-En)+2gram}  & 34.2 & \textbf{12.3} & 28.2 & \textbf{23.1} \\
        \midrule
        \textbf{Sv (Bn-En)} & \textbf{15.2}    & 17.7  & 28.2 & 20.2 \\
     
         \textbf{Ev (Bn-En)}  & 20.4   &  16.8 & \textbf{27.7} & 22.8 \\ 
     
         \textbf{Ev (Bn-En)+2gram}  & 18.4  &  \textbf{16.1} & 28.2 & \textbf{19.9} \\
         
      \bottomrule
    \end{tabular}}
  \end{center}
  \vspace{-0.3cm}
\end{table}
\begin{table}[!ht]
  \begin{center}
    \caption{WER\% results after fine-tuning the best models from \ref{table5:pre-training} on the Hi-En and Bn-En code-switching development sets.}
    \label{table7:finetuning_results}
    \resizebox{0.46\textwidth}{!}{%
    \begin{tabular}{|l|c|c|c|c|}
      \toprule 
        & \textbf{Ev(Hi-En)} &\textbf{Ev(Hi-En)+2gram } & \textbf{Ev(Bn-En)} &\textbf{Ev(Bn-En)+2gram}
      \\
       \midrule
      Hi-En CS  & 28.7 & \textbf{28.1} & - & - 
       \\
       \midrule
       Bn-En CS & - & -& \textbf{24.6} & 25.9  \\
      \bottomrule
    \end{tabular}}
  \end{center}
  \vspace{-0.3cm}
\end{table}
\begin{figure}[!ht]
\centering
 \includegraphics[width=\linewidth]{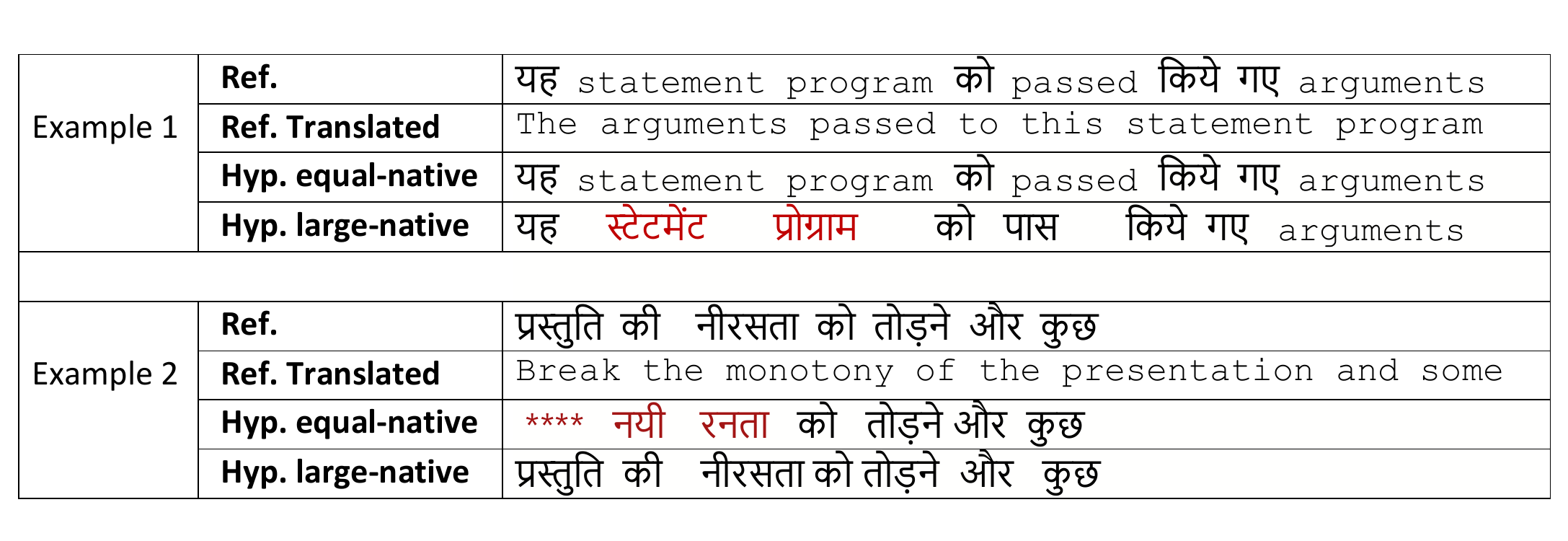}
 \vspace{-0.3cm}
  \caption{Examples from Hindi-English CS-Dev set decoded by E2E conformer model pre-trained with Ev and Sv configurations.}\label{fig2:pre-train}
  \vspace{-0.3cm}
\end{figure}
In Figure \ref{fig2:pre-train}, we show an example of decoded outputs produced from the \textit{Ev} and \textit{Sv} configurations. 
It can be noted from Example 1 that the transcription from pre-training with Ev configuration is better in identifying English words than the Sv configuration, which is more biased to Hindi characters. This is expected as, with the Sv configuration, the Hindi speech is $4$ times the amount of English speech. 
We note here that both transcriptions are phonetically correct. In addition, pre-training with Ev configuration resulted in a more robust model to misalignment since the size of the well-aligned pre-training data is larger than the Sv configuration. On the other hand, from Example $2$, we can see that pre-training with Sv produces more accurate predictions in the Hindi language.
This lead us to hypothesize that pre-training the model on a well aligned and accurate script from monolingual data resulted in robustness against inaccurate segment alignments and incorrect reference transcription presented in the CS training data. Finally, re-scoring with 2gram LM model corrects some spellings and helps better selecting the characters set for the corresponding language  as shown in Figure \ref{fig2:ngram}. However, we noticed that in some examples the LM re-scoring introduced more deletions. 
\begin{figure}[!ht]
\centering
 \includegraphics[width=\linewidth]{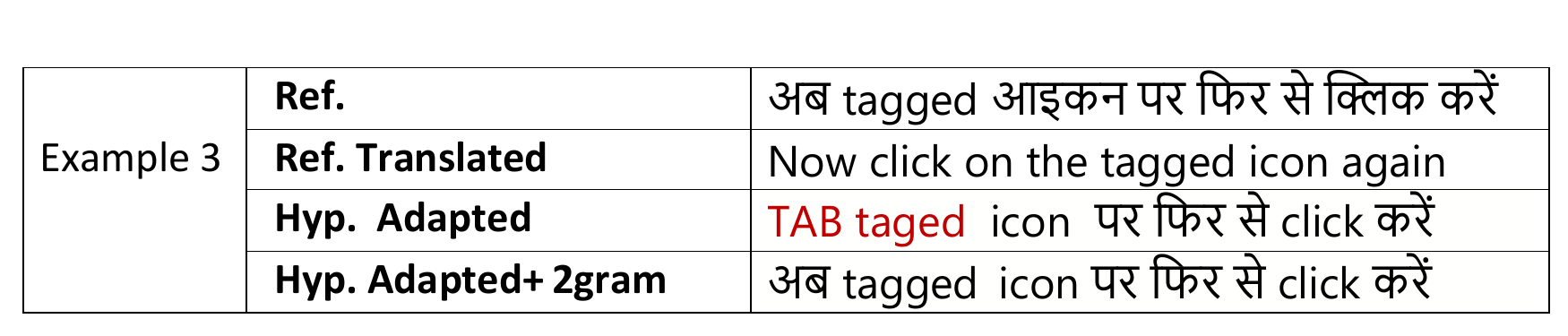}
 \vspace{-0.3cm}
  \caption{Examples from Hindi-English CS-Dev set decoded by models pre-trained with Equal non-native (Ev) configuration.}\label{fig2:ngram}
  \vspace{-0.3cm}
\end{figure}

Results on development sets, after fine-tuning is presented in Table \ref{table7:finetuning_results} shows an slight increase in the WER\% after fine-tuning on the CS data which confirms our manual observation that the CS development set is unreliable. 
But to really see the affect, we decided to report model fine-tuning results obtained from the final blind submissions. In the challenge, the systems were evaluated using the conventional word error rate (WER) and the transliterated WER (T-WER) as shown in Table \ref{table6:final_results}. 

\begin{table}[!ht]
  \begin{center}
    \caption{WER\% \& Transliterated WER (T-WER)\% results on Hi-En and Bi-En final blind set.}
    \label{table6:final_results}
    \resizebox{0.46\textwidth}{!}{%
    \begin{tabular}{|l|c|c|c|c|c|c|}
      \toprule 
        &
       \multicolumn{2}{c|}{\textbf{Hi-En}} & \multicolumn{2}{c|}{\textbf{Bn-En}} & \multirow{2}{*}{AVG WER} & \multirow{2}{*}{AVG T-WER} \\
      \cline{2-5}
       & WER & T-WER & WER & T-WER &  & 
      \\
       \midrule
       Baseline \cite{diwan2021multilingual} & 25.5 & 23.8 & 32.8 & 31.7 & 29.2 & 27.7
       \\
       \midrule
       Sv (adapted) & 23.1 & 21.2 & 27.6 & 26.3 & 25.3 & 23.7 \\
       \midrule 
       Ev (adapted) & \textbf{21.9} & \textbf{20.3} & \textbf{25.8} & \textbf{24.5} & \textbf{23.8} & \textbf{22.4}\\
      \bottomrule
    \end{tabular}}
  \end{center}
\end{table}

The T-WER counts an English word in the reference text as being correctly predicted if it is in English or in a transliterated form in the native script. It can be seen that the EV finetuned (adapted) configuration resulted in best results which confirms our findings from the pre-training phase. The rescoring with LM model corrected some mistakes. However, it also introduced some deletions. Due to the limited number of submissions, we did not consider the system with LM for final submission.

\subsection{Practical considerations for transfer learning with E2E conformer (monolingual to CS)} 
\textbf{Bilingual conformer pre-training for CS}: The monolingual pre-training for CS is very sensitive and can easily be biased to one language character set due to the phonetic overlap between the two languages. Hence, for successful pre-training for CS task, we recommend choosing the percentage of each monolingual data close to their expected percentage in the CS data. \\
\textbf{Language modeling for ASR rescoring}: Our results suggests that 2-gram model provided the best re-scoring for E2E conformer ASR in the CS scenario compared to other deep learning techniques. The rescoring with LM corrects words spelling and helps choosing the correct language character set, however, sometimes it introduces deletions.\\
\textbf{Conformer fine-tuning}: We found that fine-tuning pre-trained conformer by following the conventional freezing approach degraded the performance. Our results suggest that freezing any block in the encoder, decoder or both (encoder$+$decoder) resulted in a worse performance. To the best of our knowledge these findings are novel compared to the conventional fine-tuning with freezing part of the network for speech recognition systems. The best results were obtained from fine-tuning the entire E2E network with a learning rate of 0.1 with Noam optimizer and no warmup steps. 

\section{Conclusions}\label{conclusion}

In this paper, we have presented and evaluated our transfer learning approach for the E2E conformer-based ASR system (KARI), designed for building low-resourced code-switching ASR systems. The two steps transfer learning showed significant improvements and robustness against segment misalignment and script inconsistencies in noisy data. 
We present some practical consideration needed for the model adaptation in a real-world scenario. We showed the effect of the percentage of each selected monolingual data for pre-training, on the CS ASR performance. In future work, we plan to explore the applicability of other transfer learning methods for CS that includes self-supervised and multi-task learning approaches.

\bibliographystyle{IEEEbib}
\bibliography{refs}

\begin{thebibliography}{10}

\bibitem{sitaram2019survey}
S.~Sitaram, K.~Chandu, S.~Rallabandi, and A.~Black,
\newblock ``A survey of code-switched speech and language processing,''
\newblock {\em arXiv preprint arXiv:1904.00784}, 2019.

\bibitem{li2013improved}
Y.~Li and P.~Fung,
\newblock ``Improved mixed language speech recognition using asymmetric
  acoustic model and language model with code-switch inversion constraints,''
\newblock in {\em ICASSP}, 2013.

\bibitem{sreeram2020exploration}
G.~Sreeram and R.~Sinha,
\newblock ``Exploration of end-to-end framework for code-switching speech
  recognition task: Challenges and enhancements,''
\newblock {\em IEEE Access}, 2020.

\bibitem{amazouz2017addressing}
Djegdjiga Amazouz, Martine Adda-Decker, and Lori Lamel,
\newblock ``Addressing code-switching in {F}rench/{A}lgerian {A}rabic speech,''
\newblock in {\em Interspeech}, 2017.

\bibitem{ali2021arabic}
A.~Ali, S.~Chowdhury, A.~Hussein, and Y.~Hifny,
\newblock ``Arabic code-switching speech recognition using monolingual data,''
\newblock {\em Interspeech}, 2021.

\bibitem{hamed2021investigations}
I.~Hamed, P.~Denisov, C.~Li, M.~Elmahdy, S.~Abdennadher, and N.g Vu,
\newblock ``Investigations on speech recognition systems for low-resource
  dialectal {A}rabic-{E}nglish code-switching speech,''
\newblock {\em Computer Speech \& Language}, p. 101278, 2021.

\bibitem{chowdhury2021towards}
S.~Chowdhury, A.~Hussein, A.~Abdelali, and A.~Ali,
\newblock ``Towards one model to rule all: Multilingual strategy for dialectal
  code-switching {A}rabic {ASR},''
\newblock {\em Interspeech}, 2021.

\bibitem{toshniwal2018multilingual}
S.~Toshniwal, T.~Sainath, R.~Weiss, B.~Li, P.~Moreno, E.~Weinstein, and K.~Rao,
\newblock ``Multilingual speech recognition with a single end-to-end model,''
\newblock in {\em ICASSP}, 2018.

\bibitem{datta2020language}
A.~Datta, B.~Ramabhadran, J.~Emond, A.~Kannan, and B.~Roark,
\newblock ``Language-agnostic multilingual modeling,''
\newblock in {\em ICASSP}, 2020.

\bibitem{luo2018towards}
Ne~Luo, Dongwei Jiang, Shuaijiang Zhao, Caixia Gong, Wei Zou, and Xiangang Li,
\newblock ``Towards end-to-end code-switching speech recognition,''
\newblock {\em arXiv preprint arXiv:1810.13091}, 2018.

\bibitem{shan2019investigating}
C.~Shan, C.~Weng, G.~Wang, D.~Su, M.~Luo, D.~Yu, and L~Xie,
\newblock ``Investigating end-to-end speech recognition for
  {M}andarin-{E}nglish code-switching,''
\newblock in {\em ICASSP}, 2019.

\bibitem{kim2017joint}
S.~Kim, T.~Hori, and S.~Watanabe,
\newblock ``Joint {CTC}-attention based end-to-end speech recognition using
  multi-task learning,''
\newblock in {\em ICASSP}, 2017.

\bibitem{zhou2020multi}
X.~Zhou, E.~Y{\i}lmaz, Y.~Long, Y.~Li, and H.~Li,
\newblock ``Multi-encoder-decoder transformer for code-switching speech
  recognition,''
\newblock {\em arXiv preprint arXiv:2006.10414}, 2020.

\bibitem{diwan2021multilingual}
A.~Diwan, R.~Vaideeswaran, S.~Shah, A.~Singh, S.~Raghavan, S.~Khare, V.~Unni,
  S.~Vyas, A.~Rajpuria, C.~Yarra, et~al.,
\newblock ``Multilingual and code-switching asr challenges for low resource
  indian languages,''
\newblock {\em arXiv preprint arXiv:2104.00235}, 2021.

\bibitem{wiesner2021training}
M.~Wiesner, M.~Sarma, A.~Arora, D.~Raj, D.~Gao, R.~Huang, S.~Preet, M.~Johnson,
  Z.~Iqbal, N.~Goel, et~al.,
\newblock ``Training hybrid models on noisy transliterated transcripts for
  code-switched speech recognition,''
\newblock {\em Interspeech}, 2021.

\bibitem{conformer}
A.~Gulati, J.~Qin, C.~Chiu, N.~Parmar, Y.~Zhang, J.~Yu, W.~Han, S.~Wang,
  Z.~Zhang, Y.~Wu, et~al.,
\newblock ``Conformer: Convolution-augmented transformer for speech
  recognition,''
\newblock {\em arXiv preprint arXiv:2005.08100}, 2020.

\bibitem{guo2020recent}
Pengcheng Guo, Florian Boyer, Xuankai Chang, Tomoki Hayashi, Yosuke Higuchi,
  Hirofumi Inaguma, Naoyuki Kamo, Chenda Li, Daniel Garcia-Romero, Jiatong Shi,
  et~al.,
\newblock ``Recent developments on espnet toolkit boosted by conformer,''
\newblock {\em arXiv preprint arXiv:2010.13956}, 2020.

\bibitem{kudo2018sentencepiece}
T.~Kudo and J.~Richardson,
\newblock ``Sentencepiece: A simple and language independent subword tokenizer
  and detokenizer for neural text processing,''
\newblock {\em arXiv preprint arXiv:1808.06226}, 2018.

\bibitem{graves2006connectionist}
A.~Graves, S.~Fern{\'a}ndez, F.~Gomez, and J.~Schmidhuber,
\newblock ``Connectionist temporal classification: labelling unsegmented
  sequence data with recurrent neural networks,''
\newblock in {\em Proceedings of the 23rd international conference on Machine
  learning}, 2006.

\bibitem{heafield2011kenlm}
K.~Heafield,
\newblock ``Kenlm: Faster and smaller language model queries,''
\newblock in {\em Proceedings of the sixth workshop on statistical machine
  translation}, 2011.

\bibitem{kjartansson-etal-sltu2018}
K.~Oddur, S.~Supheakmungkol, P.~Knot, J.~Martin, and H.~Linne,
\newblock ``{Crowd-Sourced Speech Corpora for {J}avanese, {S}undanese,
  {S}inhala, {N}epali, and {B}angladeshi {B}engali},''
\newblock in {\em Spoken Language Technologies for Under-Resourced Languages
  (SLTU)}, 2018.

\bibitem{hernandez2018ted}
F.~Hernandez, V.~Nguyen, S.~Ghannay, N.~Tomashenko, and Y.~Est{\`e}ve,
\newblock ``{TED-LIUM} 3: twice as much data and corpus repartition for
  experiments on speaker adaptation,''
\newblock in {\em International Conference on Speech and Computer}, 2018.

\bibitem{ko2015audio}
Tom Ko, Vijayaditya Peddinti, Daniel Povey, and Sanjeev Khudanpur,
\newblock ``Audio augmentation for speech recognition,''
\newblock in {\em Sixteenth annual conference of the international speech
  communication association}, 2015.

\bibitem{ghahremani2014pitch}
P.~Ghahremani, B.~BabaAli, D.~Povey, K.~Riedhammer, J.~Trmal, and S.~Khudanpur,
\newblock ``A pitch extraction algorithm tuned for automatic speech
  recognition,''
\newblock in {\em ICASSP}, 2014.

\bibitem{park2019specaugment}
D.~Park, W.~Chan, Y.~Zhang, C.~Chiu, B.~Zoph, E.~Cubuk, and Q.~Le,
\newblock ``Specaugment: A simple data augmentation method for automatic speech
  recognition,''
\newblock {\em arXiv preprint arXiv:1904.08779}, 2019.

\bibitem{vaswani2017attention}
A.~Vaswani, N.~Shazeer, N.~Parmar, J.~Uszkoreit, L.~Jones, A.~Gomez, {\L}u.
  Kaiser, and I.~Polosukhin,
\newblock ``Attention is all you need,''
\newblock in {\em Advances in neural information processing systems}, 2017, pp.
  5998--6008.

\end{thebibliography}

\end{document}